\documentclass[preprint,12pt, a4paper]{elsarticle}

\usepackage{amssymb}
\usepackage{hyperref}
\journal{SoftwareX}

\begin{document}
\renewcommand{\labelenumii}{\arabic{enumi}.\arabic{enumii}}

\begin{frontmatter}

\title{NeuroAdaptTrainer: A Fiji/ImageJ Plugin for YOLO-Based Neuron Segmentation, Interactive Correction and Transfer Learning}

\author[label1]{Daniela Eraso-Casas}
\author[label2,label3]{Gerard Villarroya-Pique}
\author[label5,label4]{Esther Serrano-Pertierra}
\author[label5,label2,label4]{M. Teresa Fernández-Sánchez}
\author[label5,label3,label6]{Antonello Novelli}
\author[label1,label3]{Angel Rio-Alvarez\corref{cor1}
\author[label2,label3]{Víctor M. González}}
\cortext[cor1]{Corresponding author}
\ead{rioangel@uniovi.es}

\address[label1]{Computer Sciences Department, University of Oviedo, Asturias, Spain}
\address[label2]{Electrical Engineering Department, University of Oviedo, Asturias, Spain}
\address[label3]{Biomedical Engineering Center (BME), University of Oviedo, Asturias, Spain}
\address[label4]{Biochemistry and Molecular Biology Department, University of Oviedo, Asturias, Spain}
\address[label5]{Institute of Biotechnology of Asturias (IUBA), University of Oviedo, Asturias, Spain}
\address[label6]{Psychology Department, University of Oviedo, Asturias, Spain}

\begin{abstract}
Neuron counting and segmentation in microscopy images of neuronal cultures is a routine and time-consuming task in neuroscience research, traditionally performed through manual inspection or semi-automatic tools. We present NeuroAdaptTrainer, an open-source Fiji/ImageJ plugin that integrates a YOLO instance-segmentation model directly into the microscopist's workflow. The plugin allows a user to run automatic neuron detection on a single image or a batch of images, manually correct the resulting detections from within Fiji, and use those corrections to adapt the model to new imaging conditions via transfer learning. A built-in external validation module allows the base and adapted models to be compared quantitatively on a held-out annotated set. NeuroAdaptTrainer lowers the barrier for non-specialist users to benefit from deep-learning-based segmentation while keeping expert supervision at the center of the workflow.
\end{abstract}


\end{frontmatter}

\section*{Required Metadata}
\label{}

\section*{Current code version}
\label{}

\begin{table}[!h]
\begin{tabular}{|l|p{6.5cm}|p{6.5cm}|}
\hline
\textbf{Nr.} & \textbf{Code metadata description} & \textbf{} \\
\hline
C1 & Current code version & v1.0 (initial public release) \\
\hline
C2 & Permanent GitHub link to code/repository used for this code version & \url{https://github.com/AI-Biomedical-Engineering/NeuroAdaptTrainer} \\
\hline
C3 & Legal Code License & GNU Affero General Public License v3.0 or later (AGPL-3.0-or-later) \\
\hline
C4 & Code versioning system used & git \\
\hline
C5 & Software code languages, tools, and services used & Java (Fiji/ImageJ plugin), Python (Ultralytics YOLO, OpenCV, NumPy, PyTorch) \\
\hline
C6 & Compilation requirements, operating environments \& dependencies & Fiji/ImageJ; Java Runtime Environment; Python 3 virtual environment with Ultralytics, OpenCV, NumPy and PyTorch (optional CUDA/MPS acceleration); installers provided for macOS and Windows \\
\hline
C7 & If available, link to developer documentation/manual & \url{https://github.com/AI-Biomedical-Engineering/NeuroAdaptTrainer/blob/main/README.md} \\
\hline
C8 & Support email for questions & rioangel@uniovi.es \\
\hline
\end{tabular}
\caption{Code metadata (mandatory)}
\label{tab:metadata}
\end{table}

\textbf{Main text}\\

\begin{enumerate}

\item \textbf{Motivation and significance}

Quantifying and characterizing neurons in microscopy images of neuronal cultures is a routine task in neuroscience research, e.g. studies of neurotoxicity, neurodegeneration or drug screening. This analysis has been traditionally performed through manual visual inspection: a researcher or technician reviews each image, identifies neuronal morphology, and annotates or counts them by hand. This process is time-consuming, tiring, subjective, and prone to inter-observer variability, and it becomes a practical bottleneck as the number of images per experiment grows.

Part of the difficulty comes from the imaging modality itself. Neurotoxicity assessment usually involves observing neuronal cultures by phase-contrast of fluorescent microscopy following staining with a vital fluorescent dye (e.g., fluorescein diacetate). While the use of fluorescent dyes makes it much easier to visualize living neurons, it compromises neuronal viability and rules out longitudinal imaging of the same culture. Observation under phase-contrast, on the other hand, is non-invasive but produces images where the different elements present in the culture, i.e., neurons, glial cells, neurites and debris, are visually much harder to tell apart (Fig.~\ref{fig:differences}). An AI model trained using one modality, and images from a specific microscope and staining protocol, often needs to be re-trained before it may produce reliable results under the other imaging modality or different conditions.

\begin{figure}[!h]
\centering
\includegraphics[width=0.85\textwidth]{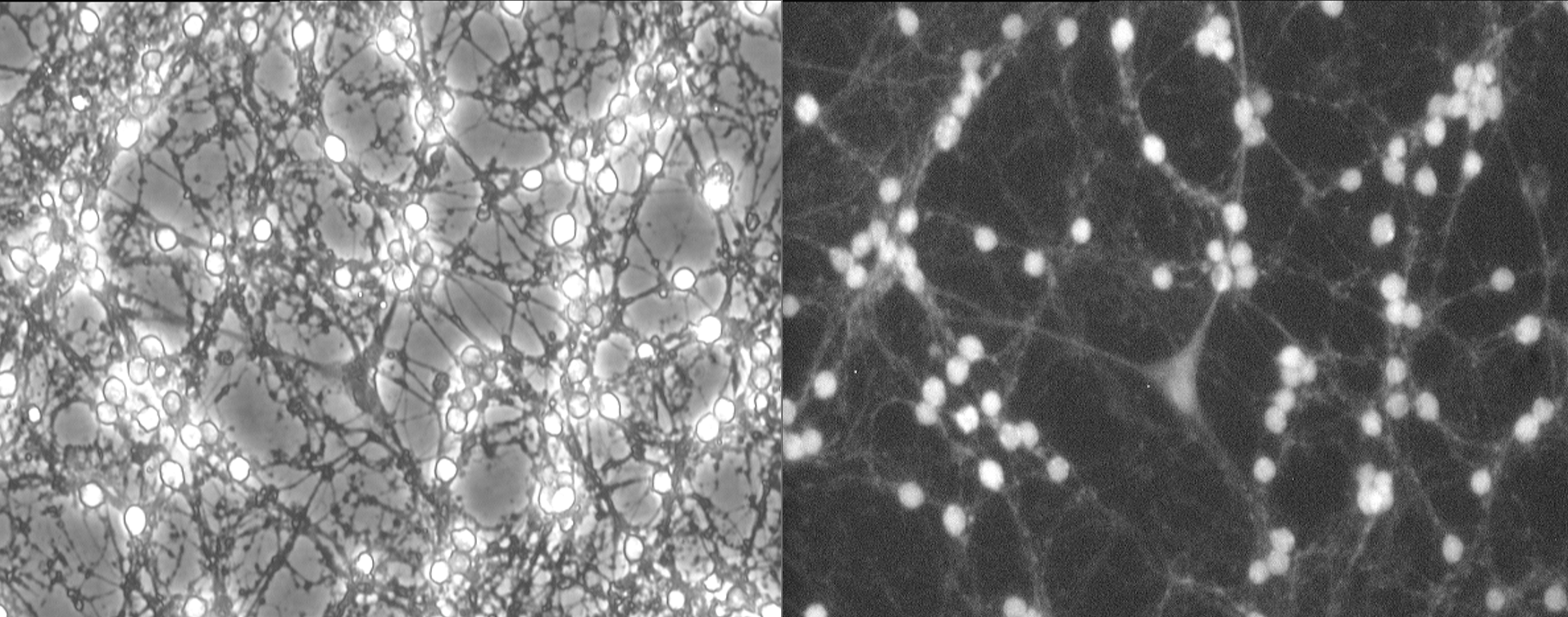}
\caption{Pictures show images of the same area of a neuronal culture using phase-contrast (left) and fluorescence (right) imaging modalities. Fluorescence improves visibility of live neurons at the cost of cell viability and longitudinal studies. Phase-contrast is non-invasive but harder to segment reliably ---illustrating why a segmentation model benefits from being adaptable across imaging conditions rather than fixed to given ones.}
\label{fig:differences}
\end{figure}

Deep-learning-based segmentation offers a way to automate this first pass, but adopting it in a wet-lab setting raises two practical problems. First, most segmentation frameworks are distributed as standalone Python scripts or notebooks, disconnected from the tools that microscopists use on a daily basis to inspect and to annotate images, namely Fiji/ImageJ~\cite{schindelin2012fiji,schneider2012imagej}. Second, models trained on one dataset frequently fail to generalize to images acquired with different microscopes, staining protocols or laboratory facilities, thus, adapting such models typically requires a machine-learning expert~\cite{pan2010transfer} who may not be available in biology laboratories. 

Existing tools partially address this gap. For example, Fiji/ImageJ provides several neuron-oriented plugins for tracing and morphometry, such as NeuronJ~\cite{imagej_neuronj} and SNT~\cite{imagej_snt}. However, these tools focus on neurite tracing rather than deep-learning-based soma detection and do not offer an in-tool re-training step. Apart from Fiji, tools such as Cellpose~\cite{stringer2021cellpose} and Ilastik~\cite{berg2019ilastik} provide general-purpose trainable cell segmentation tools. ViNe-Seg~\cite{ruffini2024vineseg} specifically combines deep-learning-assisted neuron segmentation with manual correction and custom-model retraining for calcium-imaging data through a standalone GUI. NeuroAdaptTrainer follows a similar detect-correct-retrain philosophy but it targets neuronal soma segmentation in both phase-contrast and fluorescence images with a YOLO-based instance-segmentation backbone~\cite{redmon2016yolo,ultralytics_docs}, and, distinctively, embeds the entire workflow within Fiji/ImageJ instead of in a separate application, so that users can use the same tool they are used to for the whole analysis pipeline.

NeuroAdaptTrainer addresses both problems by embedding a YOLO instance-segmentation model in Fiji/ImageJ, so that a user can obtain an automatic segmentation without using a different analysis environment. Detections are presented as editable graphical elements: false positives can be removed; missed neurons, i.e., false negatives, can be added; and boundaries can be adjusted directly on the image. These manual corrections are not discarded after being produced, instead, they are converted into YOLO-format annotations that can be used to fine-tune (transfer-learning) the current model on the user's own images, without requiring any coding from the user, therefore, generating a new model. Then, a dedicated module based on an external-validation technique can be used to compare the previous and the new models, on a held-out annotated set, so that the benefit of the adaptation step can be quantified rather than assumed. The new trained model can be adopted as the current model for further image inferences or can be discarded.

This tool does not aim to replace expert judgment or to serve as a general-purpose bio-image analysis platform; the scope of the tool is deliberately narrow: assisted detection, correction and adaptation of a YOLO segmentation model for neuronal somas in microscopy images (both phase-contrast and fluorescence), integrated in a tool that biologists are used to.

\item \textbf{Software description}

\begin{enumerate}
    \item \textit{Software architecture.} NeuroAdaptTrainer is organized in two cooperating layers: (i) \textit{a Java plugin} for Fiji/ImageJ responsible for the graphical interface, the image handling and visualization, the correction of detections, and the orchestration of the overall workflow; and (ii) \textit{a set of independent Python scripts} built on the Ultralytics implementation of YOLO, to handle the model inference, the transfer learning and the validation. The two layers communicate through several intermediate files (e.g., images, CSV files describing detections, and a configuration file) as can be seen in \ref{fig:architecture}. Such architecture keeps the plugin decoupled from the specificities of the Python/deep-learning environment, and allowing the underlying model and scripts to be updated independently of the Java interface.
    \item \textit{Software functionalities.} The plugin exposes three main windows, corresponding to three usage modes: (i) \emph{single-image segmentation}, for running and visually reviewing the model's output on one image at a time; (ii) a \emph{transfer-learning assistant}, for processing a folder of images, correcting the batch of detections, saving the corrections as reusable annotations, and launching the current model fine-tuning process; and (iii) \emph{model comparison/validation}, for evaluating a base model and a fine-tuned model on an independent annotated dataset, reporting precision, recall, mAP50 and mAP50-95 for both of them in a side-by-side layout. The plugin manages per-session temporary directories, so multiple images or folders can be processed without interference between runs.
    \item \textit{Transfer learning mechanism.} Neuronal culture images vary substantially in appearance depending on the microscope, optics, staining protocol and culture morphology. A model trained on one configuration frequently underperforms when applied to different one. Training a new model from scratch for every new imaging setup is impractical, since it would require large amounts of freshly annotated data, and considerable computational cost. NeuroAdaptTrainer instead treats adaptation as a fine-tuning problem: the currently active YOLO model is used as the starting point, and it is further trained using only the images and annotations produced during the user's own review process. Thus, a considerable small user-specific correction set is enough to adapt the model to a new optical setup or culture type. The specific workflow will be: (1) the current model is run over a folder of images; (2) the user reviews and corrects the resulting detections from within Fiji; (3) the corrections are automatically converted into a YOLO-format annotation set; (4) the fine-tuning script (built on Ultralytics YOLO) trains a new model from that set of images and user corrections, starting with the current model's weights rather than with randomly initialized ones; and (5) the resulting adapted model can be selected as the new current model for subsequent inference. Because fine-tuning is more resource-intensive than single-image inference, it is treated as an explicit, on-demand step rather than something launched automatically after every correction, and it can optionally use hardware acceleration (CUDA on compatible NVIDIA GPUs, MPS on compatible MacOS systems) when available, falling back to CPU execution otherwise. The tool does not assume that every fine-tuning run necessarily improves performance: a validation window is provided for that purpose, so that a current model and afine-tunned one can be compared on a held-out annotated set before the adapted model is trusted for further analysis, closing the detect-correct-retrain-detect loop with an explicit verification step rather than an automatic substitution of models.
    \item \textit{Sample code snippets analysis.} The Java plugin communicates with the Python back end by invoking each script as a sub-process with a fixed positional-argument convention, rather than through a network API or an embedded interpreter. For single-image inference, the plugin builds and runs a command equivalent to:
    {\footnotesize
    \begin{verbatim}
    python -u infer_one.py <input_image> <output_image> \
        <model_path> <device>
    \end{verbatim}
    }
    where \texttt{<device>} is one of \texttt{cpu}, \texttt{cuda}, \texttt{mps} or \texttt{auto\_acceleration} (the plugin lets the user choose between forcing CPU execution and requesting automatic hardware acceleration). The script's standard output is streamed back and parsed by the plugin---e.g., to report the resulting neuron count---while its standard error is redirected into the same stream and shown in the Fiji log console, so that failures in the Python environment (missing model, incompatible device,...) are visible to the user from within Fiji's environment. The transfer-learning and validation windows follow the same sub-process pattern, invoking \texttt{retrain\_model.py} and \texttt{compare\_models.py} respectively with their own fixed argument conventions. This simple, dependency-free communication mechanism is what allows the Java and Python parts of the project to be versioned, tested and updated independently of one another.
    \item \textit{Base detection model.} The YOLO instance-segmentation model bundled with NeuroAdaptTrainer (\texttt{best.pt}) was trained on an expanded, expert-level annotated dataset of neuronal cultures. Rather than relying solely on manual expert annotation, most of the training masks were produced automatically: a classical computer-vision algorithm, developed by the same team, generates neuron segmentation masks directly from fluorescence microscopy images by combining brightness-based detection with shape and size filtering, closely matching expert-drawn annotations. This automated mask-generation step facilitates to build a training set far larger than manual annotation alone would have allowed, which in turn was used to train the YOLO model integrated in the plugin. On the held-out evaluation set, the selected base model achieves a mask mAP50 of 0.9228, mask mAP50-95 of 0.4075, mask recall of 0.8970 and mask precision of 0.8707, using standard Ultralytics YOLO segmentation metrics.
\end{enumerate}

\item \textbf{Illustrative example}

A laboratory that routinely takes images of neuronal cultures under phase-contrast microscopy for drug screening purposes needs to quantify the number of live neurons across a batch of images without manually annotating each of them.

The user opens the transfer-learning assistant and selects the folder containing the batch of images as shown in Fig.~\ref{fig:segmentation}. NeuroAdaptTrainer runs the currently active YOLO model on every image and overlays the detected neuron somas directly on each image, together with a structured and editable list of detections.

\begin{figure}[!h]
\centering
\includegraphics[width=0.85\textwidth]{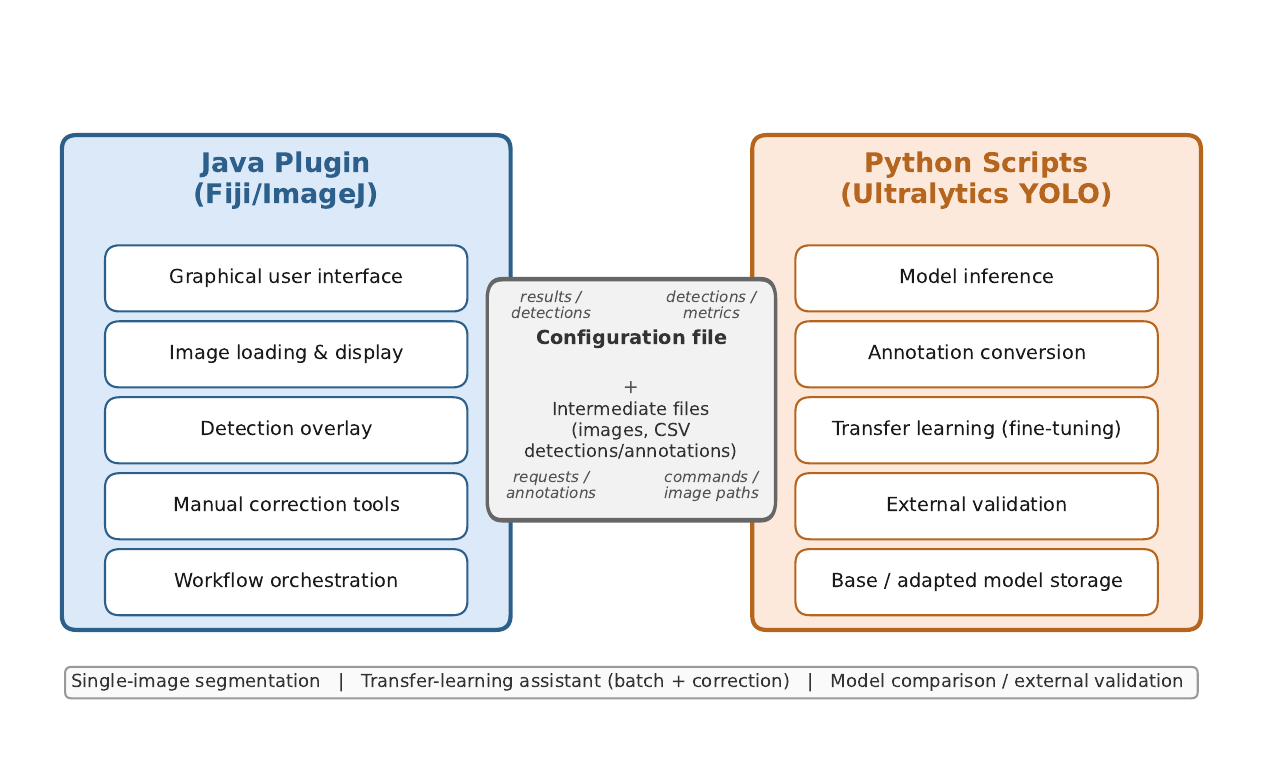}
\caption{NeuroAdaptTrainer architecture: the Fiji/ImageJ plugin (interface, review and correction) and the Python back-end (inference, transfer learning and validation), communicated through intermediate files.}
\label{fig:architecture}
\end{figure}

\begin{figure}[!h]
\centering
\includegraphics[width=0.85\textwidth]{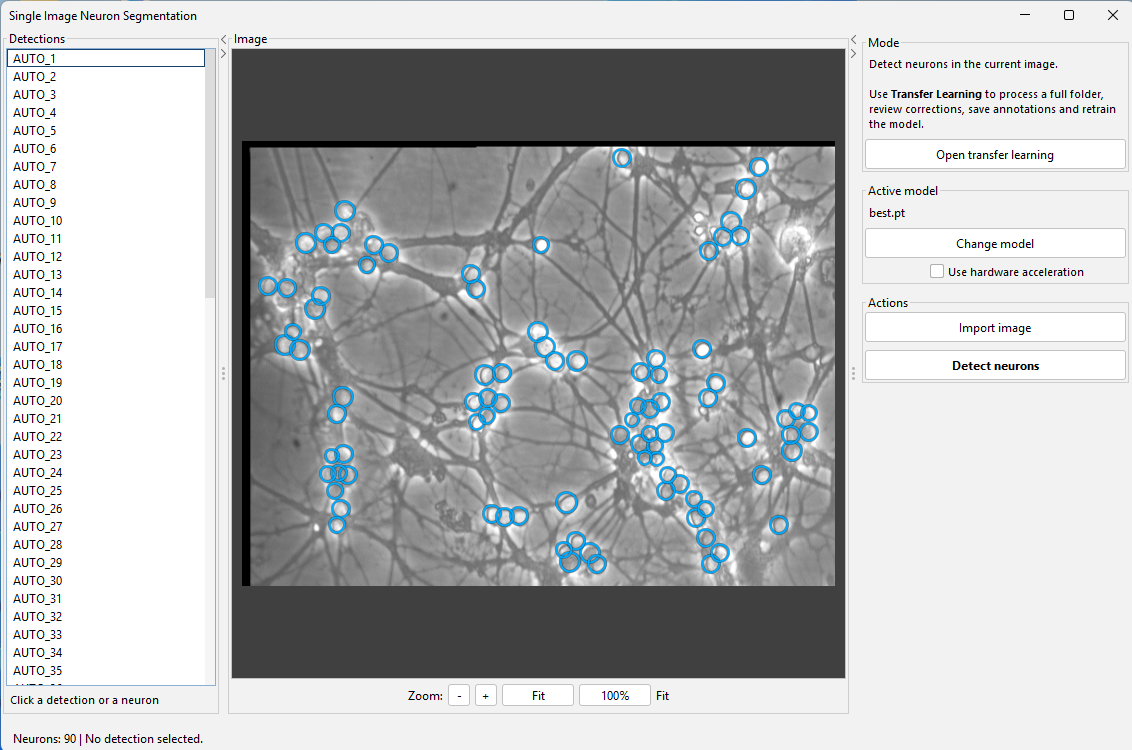}
\caption{Segmentation window: automatic YOLO detections overlaid on a neuronal culture image.}
\label{fig:segmentation}
\end{figure}

The user then reviews the batch: detections that do not correspond to real neurons (i.e., false positives) such as  glial cells or debris misidentified as neurons by the current model can easily be removed with a click; neurons missed, i.e., false negatives, by the model---which are more frequent when the imaging conditions differ from those the model was originally trained on---are added manually by outlining them on the image (Fig.~\ref{fig:correction} illustrates this correction step). All user edits update the visual overlay and the annotation list in real time, and can be saved and resumed across sessions.

\begin{figure}[!h]
\centering
\includegraphics[width=0.85\textwidth]{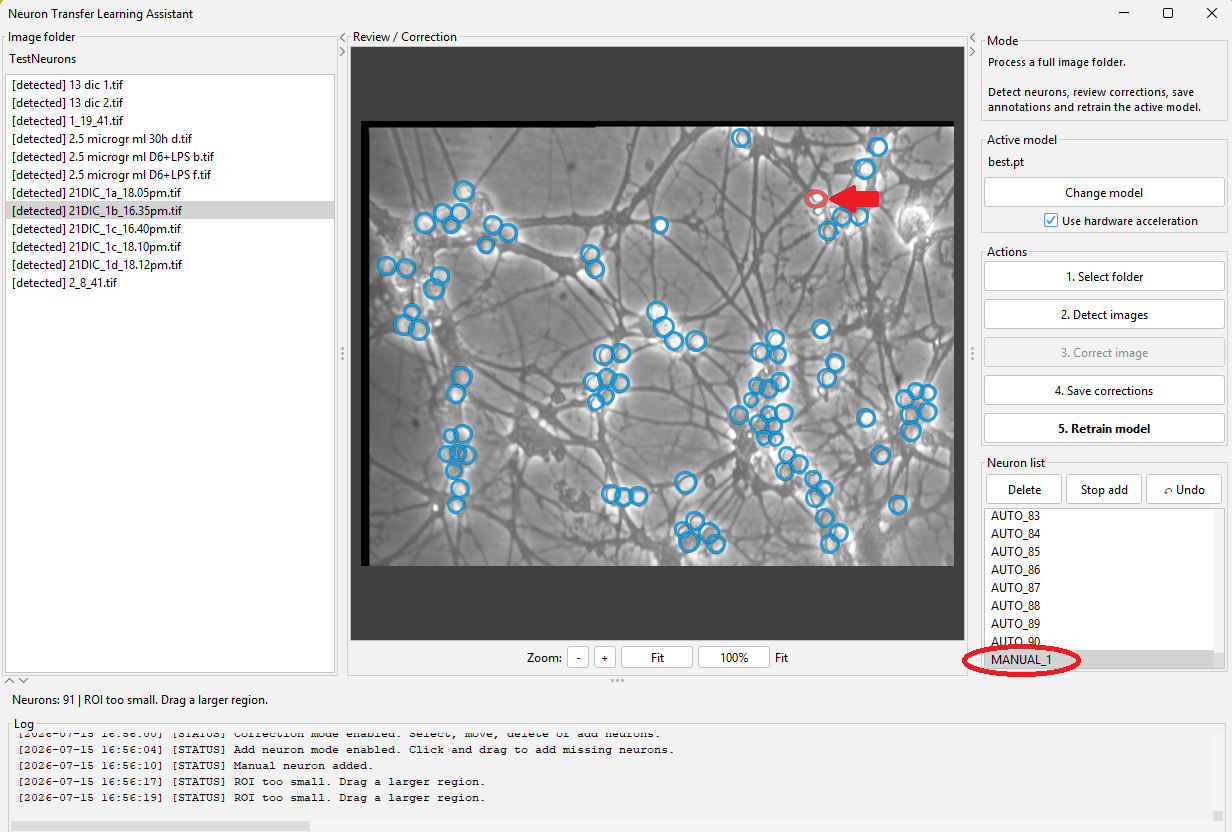}
\caption{Manual correction step: a false positive removed and a missed neuron added by the user before the corrections are saved as reusable annotations.}
\label{fig:correction}
\end{figure}

Once the user has revied the batch, he can launche the transfer-learning step from within the same window. NeuroAdaptTrainer converts the corrected annotations into a YOLO-format training set and fine-tunes the current model in the background, without the user writing a single line of code, editing configuration files, or quitting Fiji/ImageJ. The resulting adapted model can then be selected as the current model for subsequent batches of images acquired under the same conditions. It can also be compared against the previous model using the validation tool (Fig.~\ref{fig:validation}) on a held-out set of manually annotated images, so that the effect of the fine-tuning process can be easily evaluated before deciding wether to adopt it as the new current model for further analysis or not.

\begin{figure}[!h]
\centering
\includegraphics[width=0.85\textwidth]{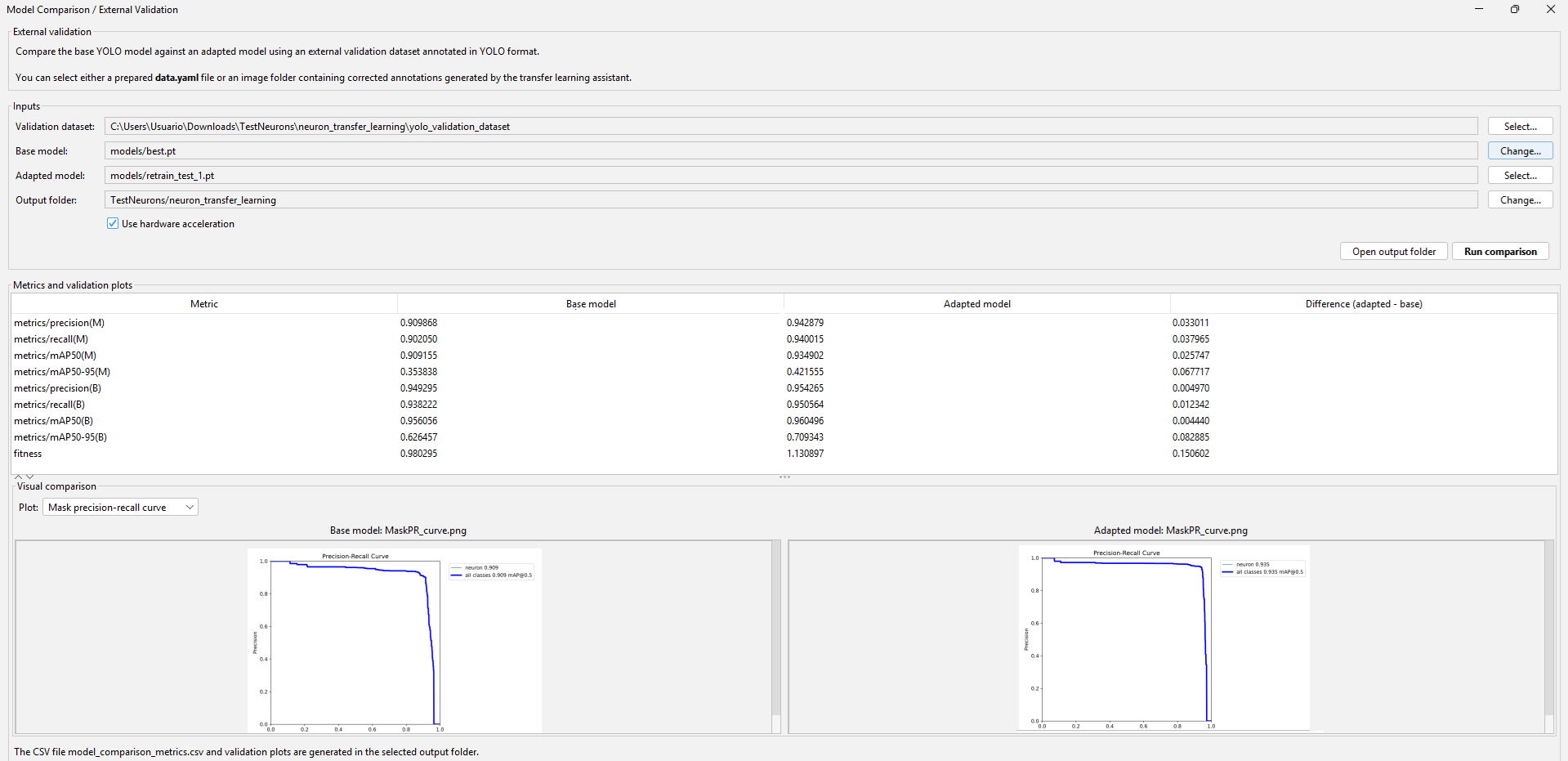}
\caption{Validation window: current (left) and adapted (right) models are compared side by side on a held-out annotated set.}
\label{fig:validation}
\end{figure}

\item \textbf{Impact}

NeuroAdaptTrainer turns a manual task normally performed with a static, pre-trained model---or not automated at all---into an interactive, adaptable workflow seamlessly integrared in the microscopist's usual tool. This allows researchers with no deep-learning expertise to (i) automatically and easily obtain segmentations of neuron somas, (ii) correct them using a familiar Fiji-based interface rather than an external annotation software, and (iii) reuse those corrections to adapt the current model to their own imaging conditions---different microscope characteristics, staining protocols, or type of culture---closing a loop that is normally split across several disconnected tools and that typically requires a machine-learning expert that may not be available in the laboratory.

This adaptability is, in our view, the main practical contribution of the tool: rather than shipping a single fixed model and hoping it will generalize to each user's images, NeuroAdaptTrainer makes model adaptation an ordinary code-free step of the analysis workflow. Beyond the specific case of neuron segmentation, the same pattern---automatic inference, in-tool manual correction, and correction-driven transfer learning, all within Fiji/ImageJ---can be applied to other instance-segmentation tasks in bio-image analysis where pre-trained models need to be fin-tuned to new image characteristics without requiring the users to code a single line or to use a different analysis environment.

\item \textbf{Conclusions}

NeuroAdaptTrainer is an open-source Fiji/ImageJ plugin that integrates YOLO-based neuron segmentation, interactive manual correction, transfer learning and model validation into a single, code-free workflow accessible to researchers lacking machine-learning background. By keeping the user in the loop at every stage---reviewing detections, correcting them, and verifying whether adaptation actually improves performance---the tool aims to make deep-learning-assisted segmentation much more trustworthy and reusable across laboratories and changing image conditions than static, non-adaptable models.

\end{enumerate}

\section*{Acknowledgements}
\label{}

This work was carried out by the AI Biomedical Engineering research group of the University of Oviedo, in the context of the bachelor thesis of Daniela Eraso-Casas under the supervision of Ángel Río-Álvarez and Víctor M. González. This work has been partially funded by the Council of Gij\'on through the University Institute of Industrial Technology of Asturias (IUTA) grants SV-25-GIJON-1-02, SV-24-GIJON-1-05, SV-22-GIJON-1-19, and SV-21-GIJON-1-19, and by Principado de Asturias, grants Severo Ochoa BP24-123 and grant AYUD/2023/36906/UO-077.

\section*{CRediT authorship contribution statement}
\label{}
\textbf{Daniela Eraso-Casas:} Software, Writing---original draft, Writing---review and editing.
\textbf{Gerard Villarroya-Pique:} Software, Writing---original draft, Writing---review and editing.
\textbf{Esther Serrano-Pertierra:} Data curation, Resources, Writing---review and editing.
\textbf{M. Teresa Fernández-Sánchez:} Data curation, Resources, Validation, Writing--review and editing.
\textbf{Antonello Novelli:} Data curation, Resources, Validation, Writing---review and editing.
\textbf{Angel Rio-Alvarez:} Software, Supervision, Conceptualization, Methodology, Project Management, Writing--original draft, Writing--review and editing.
\textbf{Víctor M. González:} Funding acquisition, Methodology, Supervision, Resources, Writing---review and editing.

\section*{Declaration of generative AI and AI-assisted technologies in the manuscript preparation process}
During the preparation of this work the authors used ChatGPT and Grammarly in order to improve the readability and language of the manuscript. After using this tool/service, the authors reviewed and edited the content as needed and take full responsibility for the content of the publication.

\section*{Current executable software version}
\label{}

\begin{table}[!h]
\begin{tabular}{|l|p{6.5cm}|p{6.5cm}|}
\hline
\textbf{Nr.} & \textbf{(Executable) software metadata description} & \textbf{} \\
\hline
S1 & Current software version & v1.0 \\
\hline
S2 & Permanent link to executables of this version & \url{https://github.com/AI-Biomedical-Engineering/NeuroAdaptTrainer/tree/main/installers} \\
\hline
S3 & Permanent link to Reproducible Capsule & \textit{Not applicable} \\
\hline
S4 & Legal Software License & GNU Affero General Public License v3.0 or later (AGPL-3.0-or-later) \\
\hline
S5 & Computing platforms/Operating Systems & macOS; Microsoft Windows; Fiji/ImageJ (cross-platform, Java-based) \\
\hline
S6 & Installation requirements \& dependencies & Fiji/ImageJ latest version; Java Runtime Environment 21+ (bundled with Fiji-latest); Python 3.10--3.12 (3.13+ not yet supported); Ultralytics YOLO, OpenCV, NumPy, PyTorch (installed automatically by the provided installers); optional CUDA-capable NVIDIA GPU or Apple Silicon (MPS) for hardware-accelerated inference and transfer learning \\
\hline
S7 & If available, link to user manual - if formally published include a reference to the publication in the reference list & \url{https://github.com/AI-Biomedical-Engineering/NeuroAdaptTrainer/blob/main/README.md} \\
\hline
S8 & Support email for questions & rioangel@uniovi.es \\
\hline
\end{tabular}
\caption{Executable software metadata}
\label{tab:executable-metadata}
\end{table}


\begin{thebibliography}{00}

\bibitem{schindelin2012fiji} Schindelin, J., Arganda-Carreras, I., Frise, E., et al. Fiji: an open-source platform for biological-image analysis. \textit{Nature Methods}, 9(7), 676--682, 2012. \url{https://doi.org/10.1038/nmeth.2019}

\bibitem{schneider2012imagej} Schneider, C. A., Rasband, W. S., Eliceiri, K. W. NIH Image to ImageJ: 25 years of image analysis. \textit{Nature Methods}, 9(7), 671--675, 2012. \url{https://doi.org/10.1038/nmeth.2089}

\bibitem{imagej_neuronj} D'Alessandro, G. Neuron Morpho plugin for ImageJ, v1.1.6. University of Southampton, 2007. \url{https://imagej.net/ij/plugins/}

\bibitem{imagej_snt} ImageJ. SNT: an ImageJ framework for tracing, visualization, quantitative analyses and modeling of neuronal morphology. \url{https://imagej.net/plugins/snt/}

\bibitem{stringer2021cellpose} Stringer, C., Wang, T., Michaelos, M., Pachitariu, M. Cellpose: a generalist algorithm for cellular segmentation. \textit{Nature Methods}, 18, 100--106, 2021. \url{https://doi.org/10.1038/s41592-020-01018-x}

\bibitem{berg2019ilastik} Berg, S., Kutra, D., Kroeger, T., et al. Ilastik: interactive machine learning for (bio)image analysis. \textit{Nature Methods}, 16, 1226--1232, 2019. \url{https://doi.org/10.1038/s41592-019-0582-9}

\bibitem{ruffini2024vineseg} Ruffini, N., Altahini, S., Weissbach, S., et al. ViNe-Seg: deep-learning-assisted segmentation of visible neurons and subsequent analysis embedded in a graphical user interface. \textit{Bioinformatics}, 40(4), btae177, 2024. \url{https://doi.org/10.1093/bioinformatics/btae177}

\bibitem{redmon2016yolo} Redmon, J., Divvala, S., Girshick, R., Farhadi, A. You Only Look Once: Unified, Real-Time Object Detection. \textit{Proc. IEEE CVPR}, 779--788, 2016. \url{https://arxiv.org/abs/1506.02640}

\bibitem{ultralytics_docs} Ultralytics. Instance Segmentation with Ultralytics YOLO. \url{https://docs.ultralytics.com/tasks/segment/}

\bibitem{pan2010transfer} Pan, S. J., Yang, Q. A Survey on Transfer Learning. \textit{IEEE Trans. Knowledge and Data Engineering}, 22(10), 1345--1359, 2010. \url{https://doi.org/10.1109/TKDE.2009.191}

\end{thebibliography}
\end{document}